%% file: main.tex
\documentclass[letterpaper, 10 pt, conference]{ieeeconf}  

\IEEEoverridecommandlockouts                              

\usepackage{graphicx} 
\usepackage{amsmath} 
\usepackage{amssymb}  
\usepackage{pseudocode}
\usepackage[ruled,vlined]{algorithm2e}
\usepackage{color}    
\usepackage[export]{adjustbox}
\usepackage[caption=true, font=footnotesize]{subfig}

\usepackage{tabularx}
\usepackage{booktabs}
\usepackage{multirow}
\usepackage{adjustbox}
\usepackage{pgfplots}
\pgfplotsset{compat=1.9}
\usepackage{lipsum}

\newcount\cyclecount \cyclecount=-1
\newcount\ind \ind=-1

\newcommand{\ie}{i.\,e.\ }

\newcommand{\eg}{e.\,g.\ }
\newcommand{\wrt}{w.\,r.\,t.\ }

\newcommand{\bfemph}[1]{\textbf{\textit{#1}}}
\definecolor{myblue}{rgb}{0.02,0.27,0.68}

\renewcommand{\baselinestretch}{0.975}

\title{
Learning Physics-Based Manipulation in Clutter: \\
Combining Image-Based Generalization and Look-Ahead Planning
} 

\author{Wissam Bejjani, Mehmet R. Dogar and Matteo Leonetti
		   \thanks{Authors are with the School of Computing, University of Leeds, United Kingdom
        {\tt\small  \{w.bejjani, m.r.dogar, m.leonetti,\}@leeds.ac.uk}}
        \thanks{This work has received funding from the UK Engineering and Physical Sciences Research Council under grant EP/R031193/1.
        }%
}

\begin{document}

\maketitle
\thispagestyle{empty}
\pagestyle{empty}


\input{Abstract}
\input{Introduction}

\input{RelatedWork}
\input{Formalism}

\input{Overview}
\input{HeuristicLearningForNovelObjects}
\input{DomainAbstraction}
\input{Experiments}

\input{realworld}
\input{Conclusion}

\bibliographystyle{IEEEtran} 
\bibliography{mybib}

\addtolength{\textheight}{-12cm}   
                                  
\end{document}

%% file: Abstract.tex
\begin{abstract}

Physics-based manipulation in clutter involves complex interaction between multiple objects.
In  this  paper,  we  consider the problem of learning, from interaction in a physics simulator, manipulation skills to  solve  this  multi-step sequential decision making problem in the real world. Our approach has two key properties: (i) the ability to  generalize and transfer manipulation skills (over the type, shape, and number of objects in the scene) using an abstract image-based representation that enables a neural network to learn useful features; and (ii) the ability to perform look-ahead planning in the image space using a physics simulator, which is essential for such multi-step problems. 
We show, in sets of simulated and real-world 
experiments (video available on \textcolor{myblue}{https://youtu.be/EmkUQfyvwkY}), that by learning to evaluate actions in an abstract image-based representation of the real world, the robot can generalize and adapt to the object shapes in challenging real-world environments. 

\end{abstract}

%% file: Introduction.tex
\section{Introduction} \label{sec:1}
\begin{figure}[!t]
\centering
\includegraphics[max width=\columnwidth]{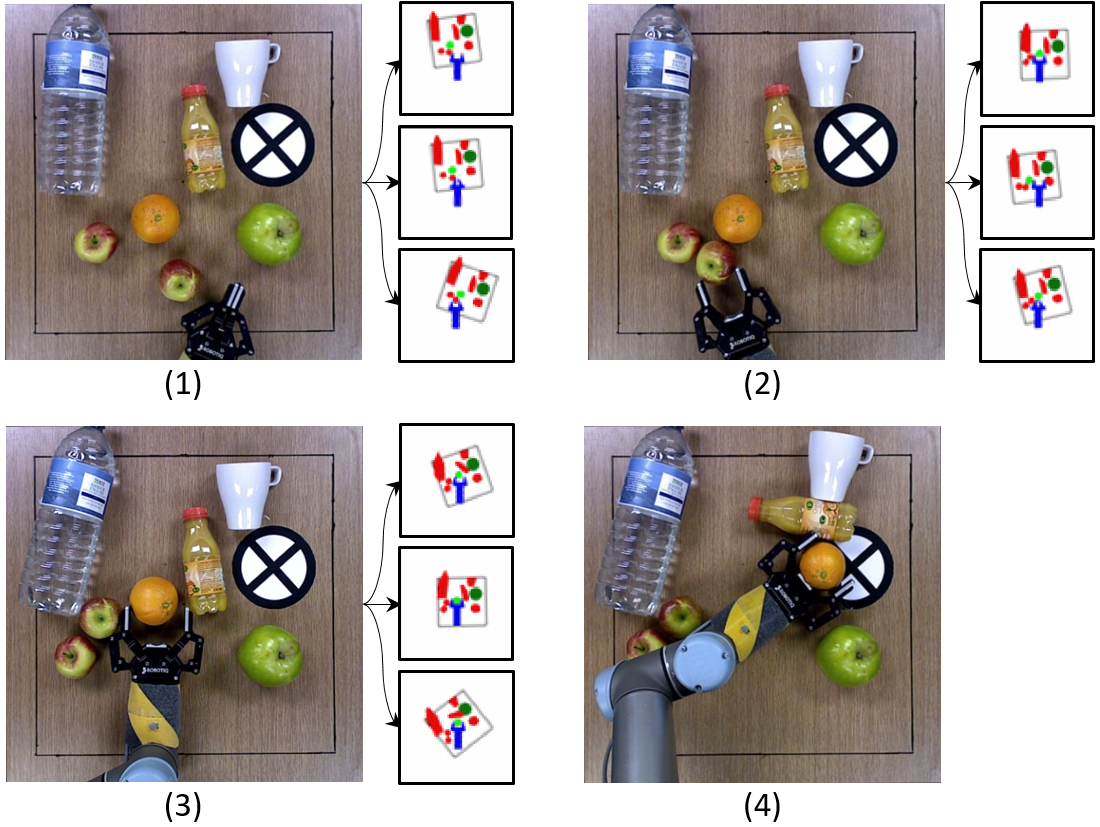}
\caption{Real-world execution of moving an object (\emph{orange fruit}) to a target location using image-based look-ahead planning.
The abstract images show the predicted horizon states under different simulated look-ahead roll-outs.
}
\label{fig:exp1}
\vspace{-4mm}
\end{figure}

The ability to acquire transferable physics-based manipulation skills is central for future robots to interact with cluttered real-world environments. Whether to pick-up items from a shelf in an industrial warehouse, or fruits from the back of a fridge, robots must execute long sequences of goal-oriented prehensile (\eg grasping) and non-prehensile (\eg pushing) manipulation actions with arbitrary objects~\cite{hernandez2016team}.
Solving such tasks requires a substantial amount of geometric and physics-based reasoning in real time. By way of illustration, consider the 
example shown in Fig.~\ref{fig:exp1}, where a robot is tasked with moving an object (the orange) to a target location, on a cluttered and constrained planar space. 
Planning a sequence of actions and executing it in open-loop will result in unintended consequences, as the interaction between the objects cannot be accurately predicted. 
The actions must, therefore, be continuously generated from a closed-loop control policy.
However, for tasks involving an arbitrary number of novel objects (not part of the training), it is not clear what features
the policy should use in order to generalize over different setups.
In this paper, we consider the problem of learning, from interaction in a physics simulator, manipulation skills that \bfemph{generalize} over everyday objects in order to solve a multi-step \bfemph{sequential} decision making problem in the real world.

Two paradigms have desirable properties, which we intend to introduce in our system: end-to-end learning, and planning-based look-ahead.
There is a momentous interest in approaching real-world manipulation tasks using end-to-end learning \cite{zeng2018learning, tobin2017domain, riedmiller2018learning}. 
End-to-end learning relieves the algorithm designer from manually having to define what features are relevant for the task. 
The problem is formulated as learning a direct mapping function from the current real-world sensory data, mainly RGB-D images, to robot joint motion.
Using images for state representation offers powerful \emph{generalization} capabilities. 



The exploration necessary for skill learning is made significantly more challenging by an unstructured state space, such as raw sensory data. This problem is exacerbated by sparse and delayed rewards.
When a model is available, solving problems with sparse reward functions can benefit enormously from incorporating look-ahead planning in the learning process, and at execution time~\cite{anthony2017thinking,  leonetti2016synthesis, leidner2018cognition}. 
This has shown to compensate for inaccuracies in the learned utility of a state-action pair~\cite{lowrey2018plan}. 
At a slightly higher computation cost, problems that require \emph{sequential} decision making in a relatively large and continuous space, can be approached in near real time.
We build on previous work~\cite{bejjani2018planning}, where a short-horizon planner was used on a physics-based simulator in conjunction with a learned value function, to both plan quickly and act robustly in a manipulation task. The value function was learned over predefined features, limiting its applicability to a given number of objects of a single particular shape.

We overcome these limitations, by proposing a novel combination of image-based learning systems with look-ahead planning, for real-world manipulation of a varying number of novel objects.
Our proposed approach relies on abstracting the real-world state to a color labelled image-based representation in a physics simulator. It allows for look-ahead planning in the image space resulting in robust manipulation skills that are transferable to different manipulation objects and environment settings.

We evaluate the proposed method and compare it to state-of-the-art approaches in a set of simulated and real-world scenarios, where the robot is faced with scenes of varying number of novel objects. 
Additionally, we show its robustness at bridging the real world with the simulated model for handling everyday objects.

%% file: RelatedWork.tex
\section{Related Work} \label{sec:2}
Most of the large  body of work on real-world manipulation in clutter addresses prehensile and non-prehensile manipulation as two separate problems.
Recent attempts have been made to combine the two~\cite{zeng2018learning, konidaris2018skills}.
Most of these attempts generate a sequence of well defined manipulation primitives, under the assumption that a manipulation task is composed of modular high level primitives, such as pushing, grasping, leveraging, or pulling.  Alternatively, other approaches have tried to blur the boundary between these primitives by, for example, combining a push and grasp motion into a single manipulation skill~\cite{dogar2012physics}.
The work of~\cite{riedmiller2018learning} proposes learning to combine tasks such as getting into contact with an object or pushing an object. Learning is performed with up to two objects in simulation, while real-world applications are demonstrated over a single object.
For what concerns tasks that require prehensile and non-prehensile manipulation actions in clutter, the work of~\cite{zeng2018learning} is the most reminiscent to ours. The system learns two distinct value functions, one for evaluating push actions and one for evaluating grasp actions. The value functions are trained on objects of predefined geometries (block objects). Consequently, they show that when everyday objects with unexpected shapes are introduced, the performance degrades significantly. 
We address this issue by mapping arbitrary objects to a common abstract state space representation.

Combining planning with Reinforcement Learning (RL) is an active field of research and is achieving many breakthrough in problems with sparse reward functions \cite{anthony2017thinking, silver2017mastering}. They mainly rely on Monte Carlo Tree Search (MCTS) to guide the RL search policy.
The vast majority of MCTS implementations uses Upper Confidence Bounds (UCB), to balance between exploitation of experienced rewards, and exploration of un-visited states. 
For these estimates to become reliable, MCTS requires running a very large number of roll-outs up to the terminal state.
The problem for adapting such a technique for manipulation in clutter is the computation cost associated with simulating the physics for state transitions. 
Indeed, it is prohibitively expensive to run this process in closed-loop (at every time step) to be of practical use in physics-based manipulation tasks.
Alternatively, approaches like Model Predictive Control (MPC), and Receding Horizon Planning (RHP), have proved more viable for practical manipulation applications. 
If the goal is not within a close reach, sequences of state-action pairs are evaluated up to a certain short horizon, then a cost function or a heuristic is used to estimate the cost-to-go from  the horizon states to the goal~\cite{agboh2018real, bejjani2018planning, lowrey2018plan}.
The approaches mentioned above assume a pre-defined set of geometric descriptions of real-world objects, and rely on their Cartesian coordinates to represent the state. 
Instead, we are interested in leveraging the object geometries 
to make the manipulation motion more efficient.

The use of images for state representation, combined with deep leaning for robot control (that is, end-to-end learning) presented a breakthrough in implicitly learning spacial features that allow for greater task generalization. 
This is attributed to the end-to-end mapping function being a Convolutional Neural Network connected to a Deep Neural Network (CNN+DNN).
Impressive implementation of end-to-end learning covers problems where the robot is tasked to grasp an object \cite{jang2017end},
push an object \cite{yuan2019end}, 
or manipulate it in hand~\cite{andrychowicz2018learning}.
Typically, the training of the system takes place on synthetic data generated in simulation. 
The data can be either collected using Imitation Learning (IL) or Deep Reinforcement Learning (DRL). 
The mapping function (CNN+DNN) must see enough variation in the data such that, at execution time, the real-world data would appear as another instance of what the network was trained on, and would thus able to generalize. 
This strategy, known as \emph{domain randomization}, has been a key element in enabling such techniques to transfer to real-world applications. 
In our work, we follow similar inspiration with the difference that we rely on abstract images rendered from the physics simulator.




%% file: Formalism.tex
\section{Problem Formulation} \label{sec:pf}
We consider the problem of manipulating, in real time, a novel object on a cluttered planner space, by means of a sequence of prehensile and non-prehensile actions. 
The robot must be able to seamlessly adapt to different geometries and clutter densities, without any of the objects falling outside of the surface boundary.  

\subsection{Formalism}
We formalize the problem as a Markov Decision Process (MDP), represented as a tuple  
$ M = \langle S,A,T,r,\gamma  \rangle $ where 
$S$ is the set of states of the environment;
$A$ is the set of actions that the robot can execute, including closing and opening the gripper;
${ T : S \times A \times S \rightarrow \mathbb{R} }$ is the transition probability function,
$r:  S \times A \rightarrow \mathbb{R} $ is the reward function;
$\gamma$ is the discount factor.

The robot interacts with the environment following a certain policy $\pi(a|s)$, then receives a reward $r(s,a)$ and the environment transitions to the next state. 
The agent's objective is to learn an approximation $\hat{q}$ of the optimal value function:
$q^*(s,a) = r(s,a) + \gamma \int_{S} T(s,a,s') \max_{a'} q(s', a') ds' $. We represent random variables with upper case letters, and their realizations with lower case letters.
From the optimal value function, it is possible to derive an optimal policy $\pi^*(a|s) = \max_a q^*(s,a)$, which, at any instant \emph{t}, maximizes the expected discounted sum of future rewards (called the \emph{return}) 
$G_t =  \sum_{k=t}^L \gamma^{k-t} R_k$, where $R_k = r(S_k,A_k)$, and $L$ is the length of the episode.
The value function $\hat{q}(s_t,a_t)$ is computed by iteratively minimizing the temporal difference error $\delta_t$:
$$\delta_t = |\hat{q}(s_t, a_t)  - y_t| $$
where the target $y_t$ is:
$$y_t = r(s_t,a_t) + \gamma \max_{a} q(s_{t+1}, a)$$

\subsection{Task Definition}\label{sec:Task_Definition}
We represent the state space as follows:
$S=\{\bold{A},\bold{V},\bold{R},\bold{G} \}$, where:
\begin{itemize}
    \item $\bold{A}$ is the arrangement of the objects : 
        \begin{equation}\label{eq:arrangement}
        \begin{aligned}
        \bold{A} = \{(x,y)^{(\textrm{desired})},(x,y)^{(2)}, \ldots, (x,y)^{(n)} \} \quad | \,
        \\
        x_{min} \leqslant  x \leqslant x_{max} 
        \quad \land \quad
        y_{min}\leqslant  y  \leqslant y_{max}
        \end{aligned}
        \end{equation}
        where $(x_{min},x_{max},y_{min},y_{max})$ forms the surface boundary
    \item $\bold{V}$ are the vertices describing the shape of objects:
        $\bold{V} = \{ v^{(\textrm{desired})}, v^{(2)}, \ldots, v^{(n)} \}$, with  $v^{(n)} \in \mathbb{R}^k$, and $k \geqslant 3$
    \item $\bold{R}$ is the Cartesian pose and gripper state of the end-effector 
        $\bold{R} = \{ ( x, y, \theta)^{(\textrm{robot})}, \theta^{(\textrm{gripper})} \}$
    \item $\bold{G}$ is the location and radius of the circular target region 
        $\bold{G}= \{ (x, y)^{(\textrm{target})}, r^{(\textrm{target})}  \quad | \quad
        x_{min} \leqslant  x \leqslant x_{max} 
        \quad \land \quad
        y_{min}\leqslant  y  \leqslant y_{max}
        \}$
\end{itemize}

We use a binary reward function, with the aim of only describing the goal, rather than favouring a particular solution through the reward: 
$$
\begin{aligned}
r(s,a) = 
\begin{cases}
0,    & \text{if} \, \ T(s,a) \in S_{\textrm{goal}} 	\\
-50,  & \text{if} \, \ T(s,a) \notin S_{\textrm{valid}} \\
-1,   & \text{otherwise}    
\end{cases}
\end{aligned}
$$
where  $S_{\textrm{valid}}$ is the set of states where all the objects are within the manipulation surface boundary; $S_{\textrm{goal}} \subset S_{\textrm{valid}}$ is the set of states where the desired object is on the target region.
The negative reward per action encourages the robot to solve the problem with as few actions as possible. 

By maximizing the reward, the robot brings the environment along a sequence of states $\langle s_t \rangle_{t=0}^{L-1} \, s.t. \,  s_t \in  S_{\textrm{valid}}$ , where $L$ is the length of the traversed states, from $s_{\textrm{init}}\in S_{\textrm{valid}}$ to $s_{\textrm{goal}}\in S_{\textrm{goal}}$. 
The arrangement $\bold{A_L}$ at a goal state is defined as:
\begin{equation}\label{eq:arrangement_T}
\begin{aligned}
\bold{A_L} = \{(x,y)^{\textrm{(desired)}}_L,(x,y)^{(2)}_L, \ldots, (x,y)^{(n)}_L\} \quad | \,
\\
x_{min} \leqslant x \leqslant x_{max} 
\quad \land \quad
y_{min}\leqslant  y  \leqslant y_{max}
\\
\land \quad 
||(x,y)^{(\textrm{desired})}_L - (x, y)^{(\textrm{target})}|| \leqslant r^{(\textrm{target})} 
\end{aligned}
\end{equation}
this corresponds to having the desired object at the target region and the the rest of the objects within the surface boundary.

%% file: Overview.tex
\begin{figure}[!t]
\centering
\includegraphics[max width=\columnwidth]{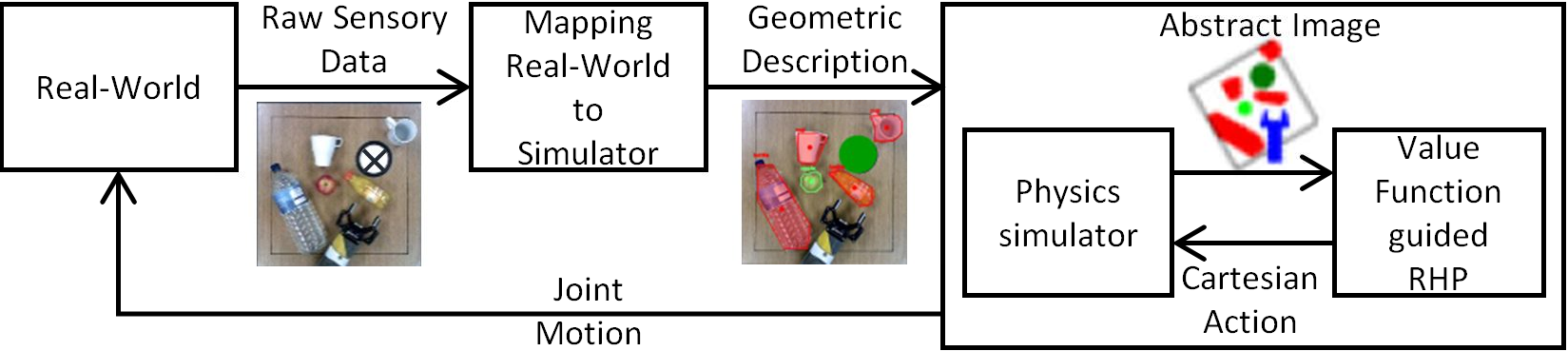}
\caption{Closed-loop control scheme for real-world execution using value function guided look-ahead planning.}
\label{fig:control}
\vspace{-5mm}
\end{figure}

\section{Overview}\label{sec:Overview}

Our proposed approach is divided into two phases: a training phase which takes place in simulation, and an execution phase which interleaves the real world with the physics simulator.

The goal of the training phase is to learn a suitable value function, in the form of a CNN+DNN, to be used as heuristic for a look-ahead planner.
We want the value function to generalize over different environment settings, namely: object shapes, clutter density, and target region location. 
To achieve generalization: (i) the data is collected in simulation over environments with different parametrizations, as detailed in Section~\ref{sec:domain_randomization} on \emph{domain randomization};
(ii) abstract images, rendered from the physics simulator, are used as state representation, to take advantage of the spacial generalization of Convolutional Neural Networks (Section \ref{sec:Sate_Representation}).
We use a Deep Reinforcement Learning (DRL) algorithm to train the CNN+DNN.
It is updated episodically from data, in the form of sequences of state-action pairs, as detailed in Section~\ref{sec:Heuristic_Learning}.
The control policy of the RL algorithm leverages its value function to guide a Receding Horizon Planner (RHP) in order to better exploit the experienced rewards, by following actions that are more likely to lead to the goal (Section~\ref{sec:RHP}).

The execution phase consists of a closed-loop control scheme illustrated in Fig.~\ref{fig:control}. 
It runs by dynamically mapping the state of the real world to the simulator, where an action is selected and then executed by the real robot.
A control scheme cycle starts by processing raw sensory data from the real world to produce a similar state in the physics simulator (Section~\ref{sec:3}).
Then, using the value function guided RHP (the same one used by the RL policy during training),
multiple roll-outs are simulated up to a certain horizon in the physics simulator. 
The state-action pairs are evaluated over the abstract images rendered from the simulator. 
Lastly, the selected action is resolved to the joint motion of the real robot.

%% file: HeuristicLearningForNovelObjects.tex
\section{Heuristic Learning For Novel Objects} \label{sec:Learning}
In this section we describe the components for learning a value function which is used to a Receding Horizon Planner (RHP). 

\subsection{Domain Randomization}\label{sec:domain_randomization}
We aim to have a system that can generalize over and, at the same time, exploit the variation 
in the shapes and number of objects, to produce an efficient behavior adapted to the particular scene.
Domain randomization has been proven effective in learning policies that can generalize over the set of randomized parameters during the training process.
The parameters that we are interested in generalizing over include: the shape and scale of the objects, the clutter density, the target location, and the initial pose of the objects and end-effector. 
We represent the parameters of a scene with the vector $\mu= 
\langle shapes,\ scales,\ clutter\ density,\\target\ location,\ initial\ distribution \rangle$. 
The shape of an object is randomly selected from a pool of polygons with random number of vertices centered around the polygon center of mass.
Then, the size of the polygon is randomly scaled up or down (within certain limits). 
This results in some of the objects being too large to fit within the gripper fingers to be grasped, whereas others are small enough to be grasped from any approach angle, and some others are directionally graspable.
Furthermore, the clutter density is also randomized by varying the number of objects in the scene.

\subsection{State Representation}\label{sec:Sate_Representation}
\begin{figure}[!t] 
    \centering
    \captionsetup[subfigure]{labelformat=empty}
    \subfloat[]{
            \adjustbox{margin=0.1em}
            {
            \begin{tabular}{c}
            Conservative\\motion
            \end{tabular}
            }
        }\hspace*{-2.0em}
    \subfloat[]{
        \includegraphics[width=0.7\columnwidth, valign=c]
        {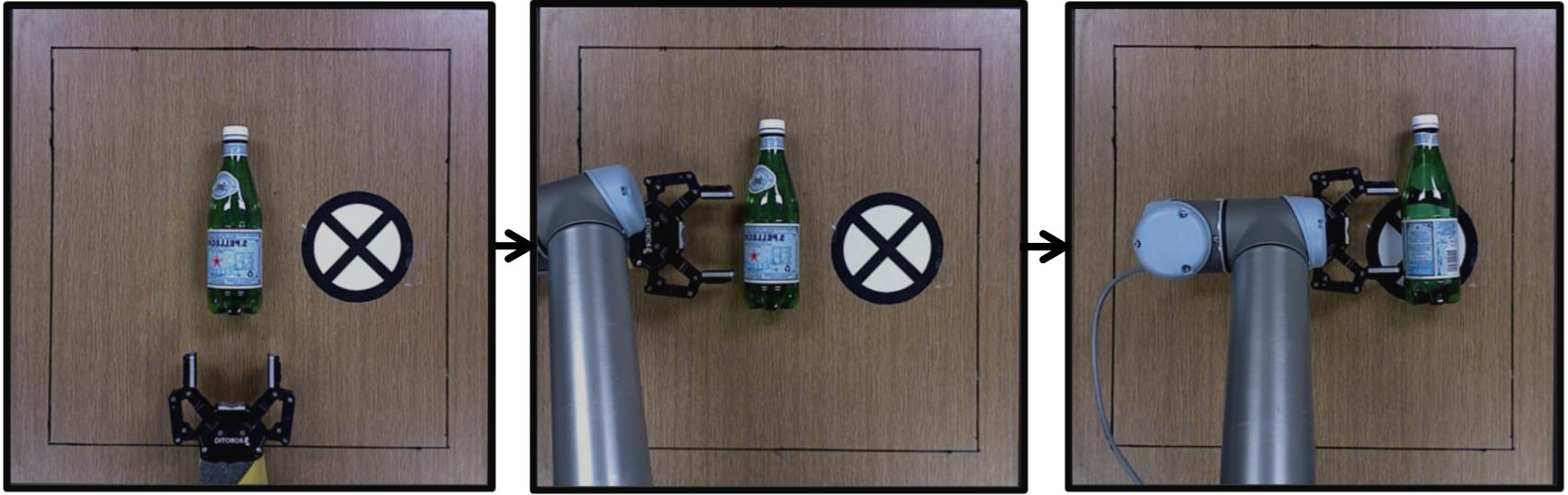}  } \\[-1ex]  
    
    \subfloat[]{
        \adjustbox{margin=1.3em}{
        \begin{tabular}{c}
        Shape\\aware\\motion
        \end{tabular}}
    }\hspace*{-1.7em}
    \subfloat[]{ \includegraphics[width=0.7\columnwidth, valign=c]{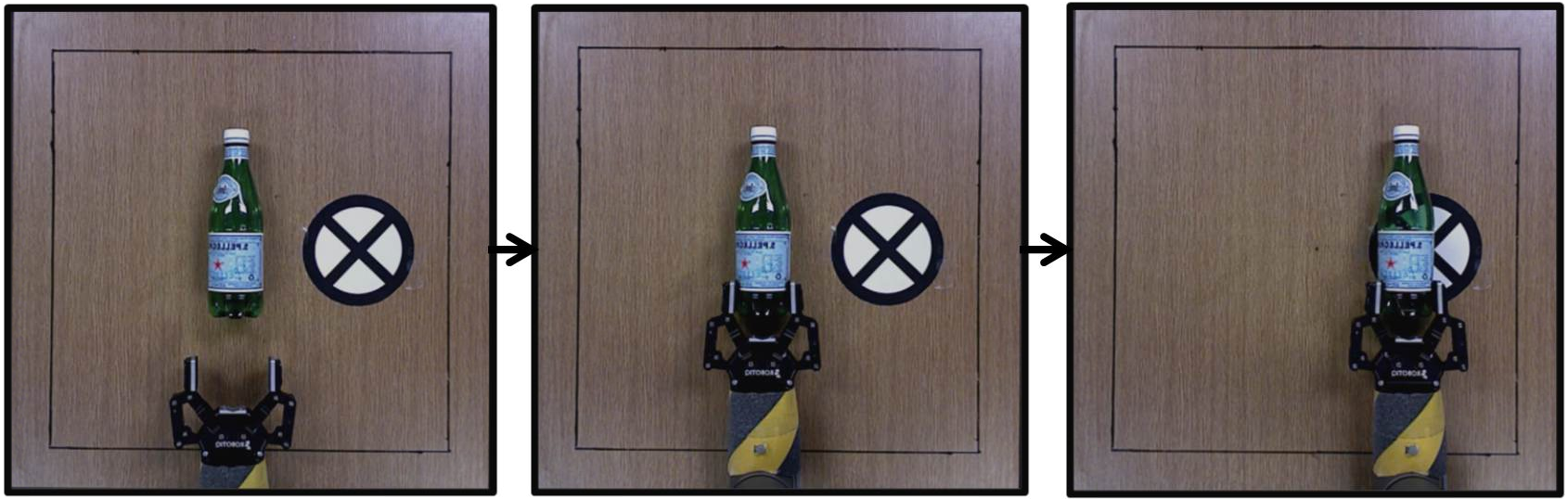}  } 
    \\[-2ex] 
    \caption{Comparing two manipulation behavior.}
    \label{fig:behavior}
    \vspace{-5mm}
\end{figure}

State representation, that is, the features on which decisions are made, play a major role in the learned behaviour.
When the parameters defining the task are not represented in the state space, the resulting policy typically converges to a robust yet conservative behavior~\cite{andrychowicz2018learning}. 
In contrast, to take advantage of the particular object shapes, those have to be available to the learner, besides the position and orientation of the objects.
For this reason, we use an image-based abstract representation of the manipulation task, where the images are rendered from the state of the physics simulator.
Such a representation makes all of the variations in the parameters $\mu$ available to the agent, and the convolutional network can capture their spatial properties to generate features for the value function.
The top example in Fig.~\ref{fig:behavior} shows a conservative behavior that works for a wide variety of object shapes, \ie to go behind the object and push it to the target region with the fingers of the gripper being in the open state.
Such a behavior can result from representations that do not encode shape features.
The bottom example of Fig.~\ref{fig:behavior} relies on image-based representation such that the robot can take advantage of the rectangular shape of the object (the bottle) by grasping it from the side and moving it to the target region. 


\subsection{Learning the Value Function}\label{sec:Heuristic_Learning}
The bulk of this work is focused on learning the CNN+DNN based value function.
It takes the abstract image-based representation of the scene, and outputs a value for the robot actions. 
The network is trained over task instances with different parameters. 
The parameters are sampled as described in~\ref{sec:domain_randomization}, and kept constant throughout the task.
We train the value function to maximize the return given the task parameters:
\begin{equation}\label{eq:valFt}
\mathbb{E} \left[ G_t \mid \mu \right].
\end{equation}
The network builds an internal representation of the features relevant for manipulating an arbitrary number of objects of different geometries. 

As mentioned in section \ref{sec:Task_Definition}, we refrained from encoding hints of the solution into the reward function, as is common in reward shaping~\cite{peng2017sim}, so as not to affect the task definition. Instead, we accelerate the learning by jump starting the value function from demonstrations~\cite{bejjani2018planning, lakshminarayanan2016reinforcement, hester2017learning}.
The demonstrations we collect are generated in the simulator using a sampling based planner, the Kinodynamic RRT~\cite{haustein2015kinodynamic}, and the rewards along the trajectory are fed to the neural network, to get an approximate value function for the policy produced by the planner. 

We then use deep Reinforcement Learning with double Q-learning and experience replay~\cite{van2016deep}, 
to optimize the value function further. 
The Reinforcement Learning algorithm starts to gradually replace the state-action pairs in the replay buffer collected from demonstrations, by the ones collected following the RL policy $\pi$ being learned.

\subsection{$\mathcal{E}$-RHP as RL Policy}\label{sec:RHP}
The exploration policy is particularly critical in DRL, especially when trained over images, since the decision is made on a vast space, not taking advantage of hand-crafted features.
For this reason, the robot must observe transition samples leading to the goal frequently enough for the value function to converge. We rely on a Receding Horizon Planner to select actions by leveraging the physics simulator within a short horizon, while learning the value function for the long-term consequences of actions, which can be more successfully approximated, since they need not to be as precise.
Therefore, we implement $\epsilon$-RHP as the RL policy. A random exploration action is selected with a probability $\epsilon$, and an RHP exploitation action is selected with probability $1-\epsilon$.

A straightforward implementation of RHP consists of running $K$ roll-outs up to a horizon $H$ \cite{lowrey2018plan, zhong2013value,bejjani2018planning}.
All roll-outs start from the current environment state $s_{\textrm{current}}$.
Each roll-out works by sampling an action $a$ according to its value with respect  to the other available actions
$$
P(a|s_t) = \frac{  exp(\hat{q}(s_t,a)/\tau)  }    {\Sigma_{a_i \in A} exp(\hat{q}(s_t,a_i)/\tau) }
$$
where $\tau$ is the temperature parameter, and $A$ is the set of available actions. 
The simulated robot is advanced along the sampled action. This process is repeated $H$ times. 
The value of a horizon state $s_H$ is also computed using the value function
$\max_{a}(\hat{q}(s_{H}, a))$.
The return of a roll-out is computed at the end of the sequence, approximating the states beyond the horizon with the current estimate of the value function:
$$
R_{0:H} =  r_1 +  \gamma r_2 + \ldots + \gamma^{H-1}r_{H} + \gamma^{H}\max_{a}(\hat{q}(s_{H}, a)).
$$
At the end of the process, the first action from the roll-out with the highest reward is executed by the robot.
Hence, the value function plays two roles in RHP. It guides the roll-outs towards promising directions, and acts as a proxy for the estimated return beyond the horizon.

%% file: DomainAbstraction.tex
\section{Mapping the Real World to the Abstract representation} \label{sec:3}
To use planning guided by the value function, in conjunction with real-world execution,
we propose mapping the state of the real world to a suitably similar state in the simulator.
Then, abstract images rendered from the simulator are used to evaluate the state-action pairs in RHP.

Our mapping focuses on the shape and functionality of the elements in the scene. 
We use real-world images of the manipulation scene and 
the robot joint configuration, to define a quantitative representation of the task
$S=\{\bold{A},\bold{V},\bold{R},\bold{G} \}$.
We apply \emph{instance segmentation} on real-world images to detect the number, location, and shape of the objects. The simulator uses this information to create polygonal objects with same contour shape as the real-world objects.
The shape of the end-effector and the dimensions of the surface are pre-loaded into the model as they do not vary from one task to another.
We use the robot Forwards Kinematics model to localize the end-effector pose in the planar Cartesian space, and the gripper state.

The input to the CNN+DNN encoding the value function is in the from of an abstract image rendered from the state of the physics simulator.  
To generate an abstract image, 
the objects in the simulator are color labeled based on their functionality.
For instance, the desired object is always of the same color, all other objects are of another common color. The same applies for the end-effector, the surface boundary, the target region, and the scene background color across all task instances.
The color labeling allows to transfer skills over different real-world setups.
For example, any object can be assigned the color of the desired object,
and the CNN+DNN will treat that object as the desired object. 

Further, the abstract images are robot centric, \ie centered around the end-effector. We found that a robot centric view reduces the learning time compared to a fixed view of the scene.

%% file: Experiments.tex
\section{Experiments} \label{sec:Exp}
We evaluate the proposed approach in series of experiments conducted in simulation and on the real robot. The goals of the experiments are threefold: 
1) to assess the effect on the performance of the main elements of the proposed approach with respect to handling a varying number of novel objects,
2) to evaluate if the acquired behavior learned to adapt to the geometries of the objects,
3) to test whether the proposed approach can
robustly control a real robot even though it uses the simulator in the control scheme.

\begin{figure}[!t]
\centering
\includegraphics[scale=0.3]{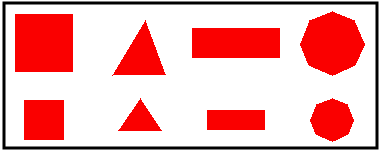}
\caption{Objects used in training process.}
\label{fig:objects}
\vspace{-6mm}
\end{figure}

We trained the value function on different task instances containing a random number of objects between $1$ and $7$.
A sample of the objects used in the training are shown in Fig.~\ref{fig:objects}.
We collected $10,000$ demonstrations using the Kinodynamic RRT planner.
We ran a total of $1,000,000$ RL episodes.
The RHP, used by the RL policy and at execution time, executes $K=4$ roll-outs of $H=4$ horizon depth each.
\subsection{Our Approach and Baseline Methods}\label{sec:bm}

We run an ablation study to assess how each element in the proposed approach affects the final performance.
Particularly, we look at the effect of the image-based abstract representation, the use of a learned heuristic, and the integration of the physics-based look-ahead planning in the control loop. Accordingly, we compose three corresponding baselines methods. All baseline methods are trained with the same procedure as ours unless otherwise specified: 
\begin{itemize}
    \item  \emph{Cartesian Pose Baseline (CaBa)}:
    Even though we trained our value function over different environment parameterizations, 
    the policy could still converge to a behavior that is robust yet impartial to the shape and density of the objects
    (\ie does the policy actually learns to adapt the behavior of the robot to the shape and number of objects in the scene). 
    Instead of using abstract images for state representation, 
    \emph{CaBa} uses the relative Cartesian poses of the objects and the target region with respect to the end-effector, and the absolute Cartesian pose of the end-effector and a binary gripper state.
    This baseline is inspired by the one used in~\cite{bejjani2018planning}.
    \item \emph{Handcrafted Heuristic Baseline (HaBa)}:
    Many planning algorithms for manipulation in clutter rely heavily on handcrafted heuristics. 
    Combined with the physics model for the local searches, we ask the question if the problem can still be solved in closed-loop without having reference to a learned heuristic \ie value function, but instead using a handcrafted heuristic to estimate the cost-to-go from a horizon state to the goal. 
    Hence, \emph{HaBa} implements RHP with a hand crafted heuristic.
    \emph{HaBa} simulate $K=15$ random roll-outs of length $H=4$.
    The heuristic used is a weighted sum of the Euclidean distances between the desired object and the target region, the rotational angle for the end-effector to face the desired object, the rotational angle for the end-effector to face the target region, and the Euclidean distances between the objects and the surface boundary. It is designed to favor a behavior where the end-effector would first approach the desired object from the back and push it towards the target region.
    \item \emph{Greedy Baseline (GreBa)}: 
    Traditionally, a RL trained agent would act greedy at execution time on the learned value function without running look-ahead planning. Albeit, in this work we started by assuming that, in an environment rich with physical interactions, it is hard for a greedy policy to anticipate the interaction dynamics.
    \emph{GreBa} challenges this claim by running a greedy policy on the value function. The value function is evaluated on the abstract image of the current state
    The simulator is, therefore, used only to generate the abstract images on which the greedy policy acts.
\end{itemize}
We note that the DNN architecture of \emph{CaBa} has an inherent limitation, which dictates that the DNN must be trained on a specific number of objects.
Adding or removing objects, \ie changing the dimension of the input space and consequently the size of the input layer, requires retraining a new DNN for the task. Hence, \emph{CaBa} uses multiple DNNs, each trained over a specific number of objects.

\subsection{Evaluation Metrics}\label{sec:em}
Data for each experiment is collected over $300$ test runs and the performance is evaluated with respect to two metrics. 
1) The \emph{Success rate} represents the percentage of the successfully completed tasks. We consider a task to be successfully completed when the desired object is moved to the target region in under $50$ actions
without having any of the objects falling off the surface edges as specified in Equation~\ref{eq:arrangement_T}.
2) The \emph{Action Efficiency} looks at how many actions were executed before successfully reaching the goal state. It is measured in view of the scene complexity which is represented by the clutter density. It is calculated as  $\frac{number\ of\ objects\ in\ the\ scene}{number\ of\ actions\ until\ completion}$. A smaller ratio implies a more conservative behavior, and a higher ratio implies a more efficient behavior that adapts to the specificity of the scene. 

\subsection{Experiments Setup}\label{sec:es}
The scenarios of the experiments consist of a number of objects laying on the surface of a table. 
The robot, a 6-DOF \emph{UR5}\footnote{https://www.universal-robots.com/products/ur5-robot/}, must use its end-effector, a \emph{Robotiq 2F-85} two finger gripper\footnote{https://robotiq.com/products/2f85-140-adaptive-robot-gripper}, to move one desired object to a target region on the table by means of prehensile and non-prehensile manipulation actions.
The surface is a square of dimension $50cm \times 50cm$. It can reasonably fit up to 7 everyday objects (ex: bottles, apples, oranges, cups, etc.). 

The real-world images are captured by a generic RGB camera.
To detect the objects and their shapes in an image, we use a pre-trained vision system, namely \emph{Mask R-CNN}~\cite{he2017mask} trained on the \emph{COCO Dataset}~\cite{lin2014microsoft}. 
The \emph{Mask R-CNN} takes a RGB image (real-world image) as an input and outputs an instance segmentation of the objects in the image.
The instance segmentation allows us to localize each object in the image together with its corresponding shape. 
The polygon' shape of an object in the simulator corresponds to the contour shapes of the object' mask outputted by the \emph{Mask R-CNN}. 
We use Box2D as our physics simulator~\cite{box2D}. 
The physics parameters (friction, inertia, and gripping force) in the simulator are empirically optimized to resemble the physics of the real world.
An abstract image is rendered from the physics simulator as follows: 
\begin{itemize}
    \item a top view image with white background centered around the end-effector
    \item the end-effector has the shape of a gripper with two articulated fingers and is colored in blue 
    \item the surface boundaries are represented by straight black lines 
    \item the desired object is colored in green
    \item all the other objects are colored in red
    \item the target region is a  circle of 5cm in radius and colored in dark green
\end{itemize}
The abstract RGB images are of $60\times60\times3$ in pixel dimension. 
The Forward and Inverse Kinematics of the UR5 are computed and simulated in OpenRAVE~\cite{diankov_thesis}.

We use the TensorFlow~\cite{tensorflow2015-whitepaper} library to build and train the Neural Networks models. 
For \emph{CaBa}, we used a feed-forward
DNN model consisting of 5 fully connected layers. 
The input corresponds to the end-effector $\bold{R}$, the objects arrangement $\bold{A}$, the target region $\bold{G}$.
The $4$ subsequent layers have $330$, $180$, $80$, and $64$ neurons, respectively, with ReLU activation functions. 
For \emph{GreBa} and \emph{our approach}, we used a CNN connected to feed-forward DNN model (CNN+DNN). 
The input is a $60$x$60$x$3$ array.  
The CNN starts with $2$ convolution layers with $32$ filters of size $6$x$6$ each, 
followed by a $2$x$2$ max-pooling layer, 
then it is followed by a sequence of convolution and $2$x$2$ max-pooling layers.
The convolution layers are $64$ $4$x$4$, $128$ $3$x$3$, respectively, with leaky ReLU activation functions. 
A flat layer connects the CNN to the DNN. 
The DNN has $4$ layers of $256$, $256$, $64$ neurons, respectively, with leaky ReLU activation functions. 
The output layer of both architectures consists of $8$ neurons with linear activation functions: $4$ for moving the robot along the cardinal directions, $2$ for rotating clockwise and counter clockwise, and $2$ for opening and closing the gripper.

\subsection{Results}\label{sec:Res}

The first set of simulated experiments examines the performance with respect to an increase in clutter density. 
The experiments consist of 4 scenarios ranging from 1 novel object on the surface (\ie no clutter, only the desired object) up to 7 novel objects. The results for the success rate and action efficiency are shown in Fig.~\ref{exp:group1SR} and Fig.~\ref{exp:group1AE} respectively.

The results show that \emph{our approach} outperforms the other baselines on both metrics.
Looking at both metrics, we observe that, for a low clutter density, all approaches present a decent level of performance. 
Not surprisingly, increasing the clutter density causes a drop in performance as it becomes much more likely for objects to fall off the edges or for the robot not to find its way through the clutter. 
We see a drastic drop in the success rate across all baselines, with 
\emph{CaBa} suffering the sharpest drop in the success rate with respect to the number of objects.
\emph{Our approach}, however, can cope better with the increase in clutter density. 
Further, all approaches show a similar increasing trend in action efficiency w.r.t. the number of objects. \emph{GreBa} and \emph{our approach} consistently hold a higher action efficiency than the other two baselines.

The second set of simulated experiments looks at the performance with respect to an increase in the size of the objects relative to the dimensions of the gripper. 
We expect that the shape of small objects is less significant to the manipulation task compared to large objects.
Specifically, one setting uses a random number of small novel objects. The small objects are chosen such they can fit inside the fingers of the gripper (\ie graspable objects). Another setting uses a random number of large novel objects. The large objects are chosen such that they cannot be grasped by the gripper. We also include a setting which has a mix of small and large objects. 
The results are reported respectively in Fig.~\ref{exp:group2SR} and Fig.~\ref{exp:group2AE} for the success rate and the action efficiency.

The results show that large objects seem to be slightly more difficult to manipulate as reflected in a decrease in the success rate. 
Albeit, \emph{our approach} appears to be significantly more robust to the increase in object sizes. 
In addition, the action efficiency shows no significant variation between the different settings, but similar to the first set of experiments, \emph{GreBa} and \emph{our approach} consistently score higher in terms of action efficiency. 

We also present the average execution and planning time in Table~\ref{tab:time}. The results are averaged over the $300$ experiments with random number of objects of novel shapes.
RHP causes a significant jump in computation time as evident by \emph{GreBa}'~low computation time, in which the policy acts greedy on the value function without running any physics roll-outs. 
The difference in time between \emph{CaBa} and \emph{our approach} is due to the difference in inference time between using the DNN and CNN$+$DNN respectively. 
We allowed for more RHP roll-outs to be simulated for \emph{HaBa} to compensate for any deficiency in the handcrafted heuristic. Therein, the average planning and execution time for \emph{HaBa} is the highest.

\begin{table}[!t]
\centering
\caption{Average Planning and Execution Time per Task.}
\begin{tabular}{lllll}
\hline
\multicolumn{1}{c|}{Method} & \multicolumn{1}{c|}{CaBa} & \multicolumn{1}{c|}{HaBA} & \multicolumn{1}{c|}{GreBa} & \multicolumn{1}{c}{Our Appr.} \\ \hline
\multicolumn{1}{c|}{\begin{tabular}[c]{@{}c@{}}Time \\ in seconds\end{tabular}} & \multicolumn{1}{c|}{19.4$\pm$2.5} & \multicolumn{1}{c|}{38.7$\pm$3.6} & \multicolumn{1}{c|}{2.4$\pm$ 0.2} & 27.1$\pm$ 2.2 \\ \hline
 &  &  &  &  \\
 &  &  &  & 
\end{tabular}
\label{tab:time}
\vspace{-9mm}
\end{table}

\subsection{Evaluation}\label{sec:Eval}
As expected, the prevalence of rich physical interactions in the environment makes the problem harder to solve. 
The fact that \emph{CaBa} scores consistently lower than \emph{GreBa} and \emph{our approach} validates the hypothesis that an expressive yet sufficient state space representation is crucial to the final performance. 
Without the geometric details of the objects, \emph{CaBa} converges to a behavior that suites the average variation in the object shapes. 
Albeit, this shows to be problematic with the high clutter density and large objects.

Further, even using a meticulously handcrafted heuristic in \emph{HaBa} and at a high computation cost it still underperforms compared to the one learned over the abstract image-based representation. This is because the learned heuristic represents a good estimate of the optimal value function whereas the handcrafted one is based on the intuition of the algorithm designer. 
We suspect that further tuning the weights that balance the handcrafted heuristic might slightly enhance its performance.

\emph{GreBa} shows decent level of performance contrary to our expectation particularly when the clutter density is not very high. 
Nevertheless, it still follows the trend of a dramatic drop in performance with an increase in clutter density. 
Hence, we conclude that having a physics model in the closed-loop control scheme is necessary to alleviate the complexity associated with anticipating the outcome of physical interactions. 

\begin{figure}[t]
\begin{tikzpicture}
  \begin{axis}[
    axis x line*=bottom,
    axis y line*=left,
    y post scale=0.35,
    ybar,
    ymin=0,
    ymax=0.3,
    ylabel={Action Efficiency}, 
    enlarge x limits=0.15,
    legend style={at={(0.09,1.1)},anchor=north west,legend columns=-1, font= \footnotesize},
    symbolic x coords={Small objects,Mixed objects,Large objects},
    xtick=data, 
    font=\small,
    ybar=3pt,
    bar width=6pt,
    x tick label style={rotate=0,align=center, text width=1.5cm, font=\small},
    ]
    \addplot+[error bars/.cd,y dir=both,y explicit, x dir=both,x explicit]
    coordinates {
    (Small objects,  0.17)     +-= (0,0.01) 
    (Mixed objects,  0.15)     +-= (0,0.02) 
    (Large objects,  0.16)     +-= (0,0.03) 
    };
    \addplot+[error bars/.cd,y dir=both,y explicit] 
    coordinates {
    (Small objects,  0.16)     +-= (0,0.02) 
    (Mixed objects,  0.16)     +-= (0,0.03) 
    (Large objects,  0.14)     +-= (0,0.01) 
    };
    \addplot+[error bars/.cd,y dir=both,y explicit]   
    coordinates {
    (Small objects,  0.22)     +-= (0,0.01)
    (Mixed objects,  0.21)     +-= (0,0.02) 
    (Large objects,  0.19)     +-= (0,0.02) 
    };
    \addplot+[error bars/.cd,y dir=both,y explicit]   
    coordinates {
    (Small objects,  0.22)     +-= (0,0.02)
    (Mixed objects,  0.22)     +-= (0,0.02) 
    (Large objects,  0.19)     +-= (0,0.02)
    };
    \legend{CaBa, HaBa, GreBa, Our Approach}
  \end{axis}
\end{tikzpicture}
\caption{Action efficiency \wrt object sizes.}
\label{exp:group2AE}
\end{figure}
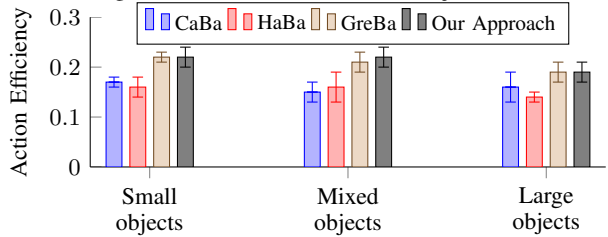

Looking at the action efficiency results in Fig.~\ref{exp:group1AE} and Fig.~\ref{exp:group2AE}, we see a clear trend where \emph{GreBa} and \emph{our approach} have similarly high action efficiency than the other baselines. 
Having in common the abstract image-based representation of the state space is strong indicator that the learned value function managed to capture relevant features from the state space representation.
This further supports our claim that the CNN+DNN is able to capture spacial features from images which allow for greater generalization and more efficient behaviors.



\begin{figure}[t]
\begin{tikzpicture}
	\begin{axis}[
    x post scale=0.95,
    y post scale=0.60,
    xmin=1,
    ymin=00,
    ymax=100,
    ylabel={Success Rate}, 
    xlabel={Number of objects}, 
    axis x line*=bottom,
    axis y line*=left,
    legend style={at={(0,0)},anchor=south west,legend columns=1, font=\footnotesize},
	]
	\addplot coordinates
		{(1,81) (3,75) (5,46) (7,26)};
	\addplot coordinates
		{(1,83) (3,72) (5,60) (7,47)};
	\addplot coordinates
		{(1,92) (3,86) (5,76) (7,60)};
	\addplot coordinates
		{(1,99) (3,96) (5,92) (7,81)};
	\legend{CaBa, HaBa, GreBa, Our Approach}
	\end{axis}
\end{tikzpicture}
\caption{Performance \wrt clutter density.}
\label{exp:group1SR}
\vspace{-2mm}
\end{figure}
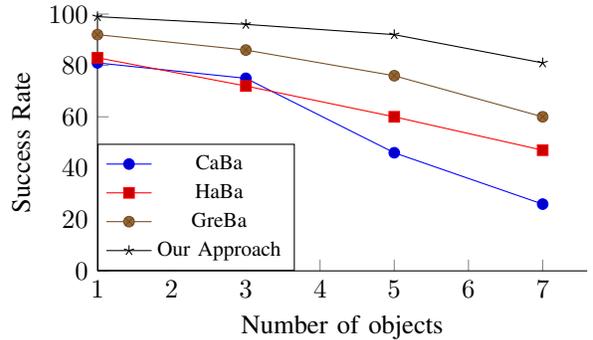

\begin{figure}[t]
\begin{tikzpicture}
  \begin{axis}[
    axis x line*=bottom,
    axis y line*=left,
    y post scale=0.35,
    ybar,
    ymin=0,
    ymax=0.6,
    ylabel={Action Efficiency}, 
    enlarge x limits=0.15,
    legend style={at={(0.09,1.1)},anchor=north west,legend columns=-1, font=\footnotesize},
    symbolic x coords={1 object,3 objects,5 objects,7 objects},
    xtick=data, 
    font=\small,
    ybar=3pt,
    bar width=6pt,
    x tick label style={rotate=0,align=center, text width=1.5cm, font=\small},
    ]
    \addplot+[error bars/.cd,y dir=both,y explicit, x dir=both,x explicit]
    coordinates {
    (1 object,   0.06)     +-= (0,0.01) 
    (3 objects,  0.16)     +-= (0,0.01)  
    (5 objects,  0.25)     +-= (0,0.05)  
    (7 objects,  0.32)     +-= (0,0.03) 
    };
    \addplot+[error bars/.cd,y dir=both,y explicit] 
    coordinates {
    (1 object,   0.04)     +-= (0,0.01)
    (3 objects,  0.14)     +-= (0,0.02) 
    (5 objects,  0.23)     +-= (0,0.02)  
    (7 objects,  0.34)     +-= (0,0.05)  
    };
    \addplot+[error bars/.cd,y dir=both,y explicit]   
    coordinates {
    (1 object,   0.07)     +-= (0,0.01) 
    (3 objects,  0.20)     +-= (0,0.01) 
    (5 objects,  0.31)     +-= (0,0.02) 
    (7 objects,  0.41)     +-= (0,0.04) 
    };
    \addplot+[error bars/.cd,y dir=both,y explicit]   
    coordinates {
    (1 object,   0.08)     +-= (0,0.01) 
    (3 objects,  0.21)     +-= (0,0.01) 
    (5 objects,  0.33)     +-= (0,0.02) 
    (7 objects,  0.42)     +-= (0,0.03) 
    };
    \legend{CaBa, HaBa, GreBa, Our Approach}
  \end{axis}
\end{tikzpicture}
\caption{Action efficiency \wrt clutter density.}
\label{exp:group1AE}
\vspace{-6mm}
\end{figure}
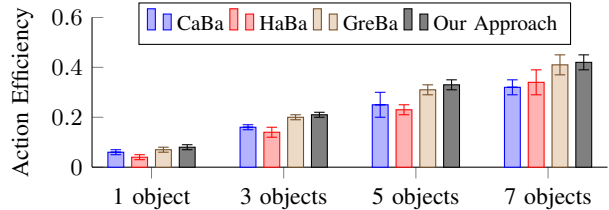

\begin{figure}[t]
\begin{tikzpicture}
	\begin{axis}[
    x post scale=0.95,
    y post scale=0.60,
    ymin=40,
    ymax=100,
    ylabel={Success Rate}, 
    axis x line*=bottom,
    axis y line*=left,
    symbolic x coords={Small objects,Mixed objects, Large objects},
    xtick=data, 
    font=\small,
    x tick label style={rotate=0,align=center, text width=1.5cm, font=\small},
    legend style={at={(0,0)},anchor=south west,legend columns=1, font=\footnotesize},
	]
	\addplot coordinates
		{(Small objects,78) (Mixed objects,65) (Large objects,60)};
	\addplot coordinates
		{(Small objects,76) (Mixed objects,70) (Large objects,65)};
	\addplot coordinates
		{(Small objects,90) (Mixed objects,81) (Large objects,74)};
	\addplot coordinates
		{(Small objects,99) (Mixed objects,95) (Large objects,92)};
	\legend{CaBa, HaBa, GreBa, Our Approach}
	\end{axis}
\end{tikzpicture}
\caption{Performance \wrt to objects sizes.}
\label{exp:group2SR}
\vspace{-5mm}
\end{figure}
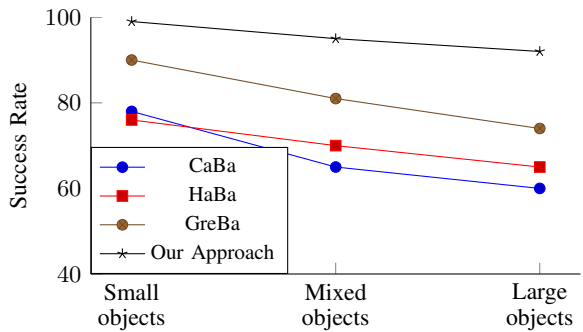

%% file: realworld.tex
\subsection{Real-World Experiments} \label{sec:realWorld}
\begin{figure}[!t]
    \centering
    \includegraphics[max width=\columnwidth]{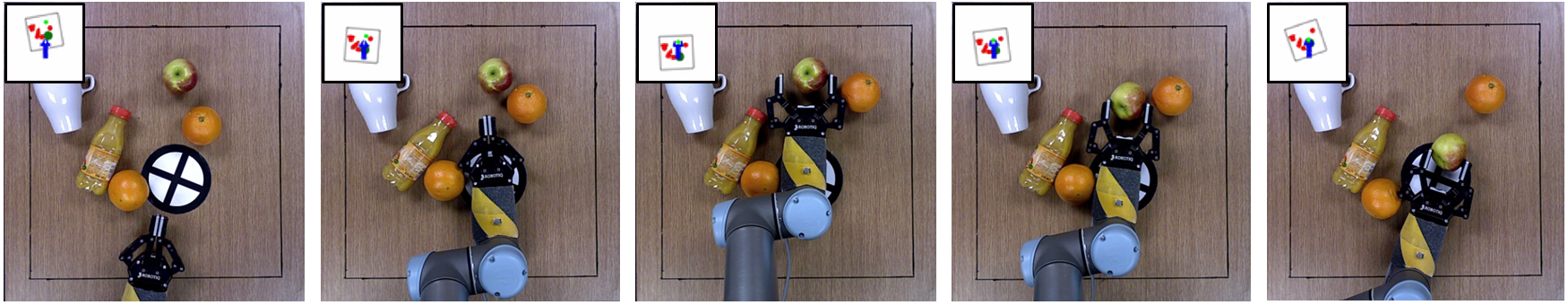}
    \caption{Moving the \emph{small apple} to the target region.}
    \label{fig:exp2}
    \vspace{-1mm}
\end{figure}

\begin{figure}[!t]
    \centering
    \includegraphics[max width=\columnwidth]{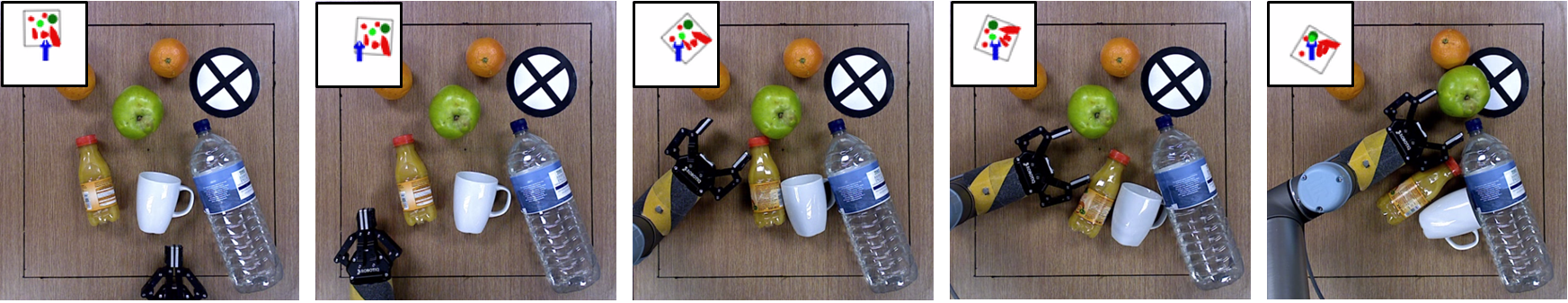}
    \caption{Moving the \emph{large apple} to the target region.}
    \label{fig:exp3}
    \vspace{-1mm}
\end{figure}

\begin{figure}[!t]
    \centering
    \includegraphics[max width=\columnwidth]{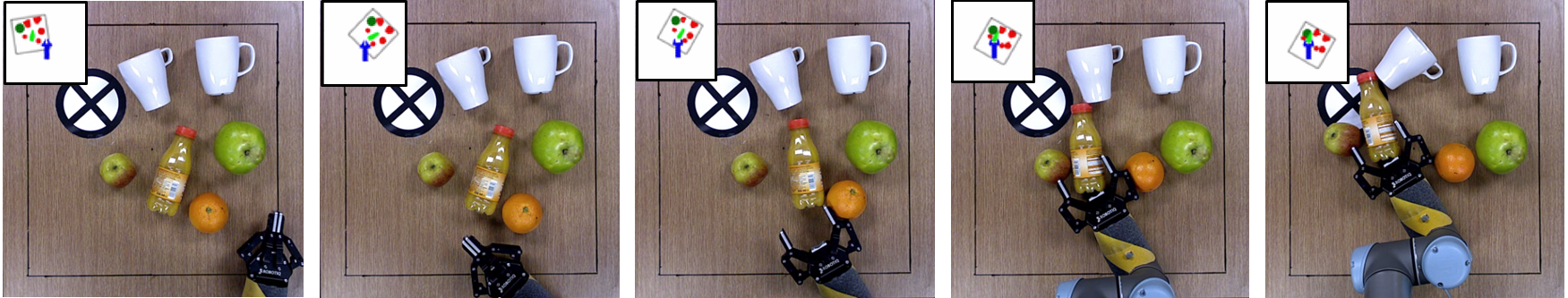}
    \caption{Moving the \emph{juice bottle} to the target region.}
    \label{fig:exp4}
    \vspace{-5mm}
\end{figure}
In this section, we reach main target of our work which is centered around handling everyday real-world objects. 
We have identified from the previous section that \emph{our approach} can handle a random number of objects of different geometries.
We run several real-world experiments to validate 
the robustness of \emph{our approach} 
in transferring manipulation skills between the simulation domain and the real world. 
The experiments involve a variety of objects of different size and shapes. For example, small and large apples, ranges, cups, and bottles.

We present snippets from some of the experiments in Fig.~\ref{fig:exp1}, \ref{fig:exp2}, \ref{fig:exp3}, and \ref{fig:exp4}. The abstract image-based representation of the environment is overlaid on the corner of the images.
We can observe an interesting behavior in Fig.~\ref{fig:exp2} where the robot is tasked to move the \emph{small apple} to the target region. The robot approaches the apple with the gripper closed, then grasp it and pull it back to the target region. 
On the other hand, in Fig. \ref{fig:exp3} where the robot has to handle a \emph{large apple}, we notice that the robot went first to the left side of the surface, pushed the clutter to the side, then went to pushing the \emph{large apple} without attempting to grasp it.
Another fascinating behavior is observed in Fig.~\ref{fig:exp4} where the robot is tasked to move the \emph{juice bottle} to the target region. First, we see the robot exploiting the rectangular shape of the bottle by maneuvering it into stable position within the fingers of the gripper, then carefully driving to the target region, all the while interacting with the clutter without causing any object to fall outside the surface boundary.
A full video of the experiments is available on \textcolor{myblue}{https://youtu.be/EmkUQfyvwkY}.

The robot was able to seamlessly transfer skills between the two domains whilst generalizing to real-world objects that were not experienced before. 
The overall behavior is robust against temporary failures in object detection and can dynamically adjust to the object dynamics.

%% file: Conclusion.tex
\section{Conclusions} \label{sec:Conclusion}
This paper described a hybrid combination of real-world execution with planning in a simulated space.
By reasoning over abstract images on the possible outcome of a manipulation motion, the control policy was able to generate, in closed-loop, complex sequences of manipulation actions.
To the best of our knowledge, this is the first work to achieve generalization over arbitrary number and shape of everyday objects in a planar manipulation task using prehensile and non-prehensile actions. 
We demonstrated the promising potential of the hybrid control scheme and its possible implication for real-world applications. 

We are building on our findings to explore how we can continuously infer and abstract arbitrary dynamic properties from the objects (\eg directional rolling, omnidirectional sliding, quasistatic, etc.) in order to exploit them for making manipulation motions even more efficient.

